\documentclass{article}

 \PassOptionsToPackage{numbers, compress}{natbib}

     \usepackage[preprint]{neurips_2019}

\usepackage[utf8]{inputenc} 
\usepackage[T1]{fontenc}    
\usepackage{hyperref}       
\usepackage{url}            
\usepackage{booktabs}       
\usepackage{amsfonts}       
\usepackage{nicefrac}       
\usepackage{microtype}      

\usepackage{float}
\usepackage{graphicx}
\usepackage{subfigure}
\usepackage[tbtags]{amsmath}
\usepackage{multicol}
\usepackage{dashrule}
\usepackage{makecell}

\title{Anti-Confusing: Region-Aware Network for Human Pose Estimation}

\author{%
  \textbf{Xuan Cao \quad Yanhao Ge \quad Ying Tai \quad Wei Zhang} \\ 
  \textbf{Jian Li \quad Chengjie Wang \quad Jilin Li \quad Feiyue Huang} \\
  Tencent Youtu Lab\\
  \texttt{\{marscao, halege, yingtai, gavinwzhang\}@tencent.com} \\
  \texttt{\{swordli, jasoncjwang, jerolinli, garyhuang\}@tencent.com} \\
}

\begin{document}

\maketitle

\begin{abstract}
    In this work, we propose a novel framework named Region-Aware Network (RANet), which learns the ability of anti-confusing in case of heavy occlusion, nearby person and symmetric appearance, for human pose estimation.
    Specifically, the proposed method addresses three key aspects, \textit{i.e.}, data augmentation, feature learning and prediction fusion, respectively.
    First, we propose Parsing-based Data Augmentation (PDA) to generate abundant data that synthesizes confusing textures.
    Second, we not only propose a Feature Pyramid Stem (FPS) to learn stronger low-level features in lower stage; but also incorporate an Effective Region Extraction (ERE) module to excavate better target-specific features.
    Third, we introduce Cascade Voting Fusion (CVF) to explicitly exclude the inferior predictions and fuse the rest effective predictions for the final pose estimation. 
    Extensive experimental results on two popular benchmarks, \textit{i.e.} MPII and LSP, demonstrate the effectiveness of our method against the state-of-the-art competitors.
    Especially on easily-confusable joints, our method makes significant improvement.
\end{abstract}

\vspace{-0.3cm}
\section{Introduction}
\begin{figure}[t!]
\setlength{\abovecaptionskip}{-0.1cm} 
\setlength{\belowcaptionskip}{-0.3cm}
\label{fig_confusing}
\centering
\subfigure[Heavy occlusion]{
\begin{minipage}[t]{0.231\linewidth}
\centering
\small
\includegraphics[width=1.0\columnwidth]{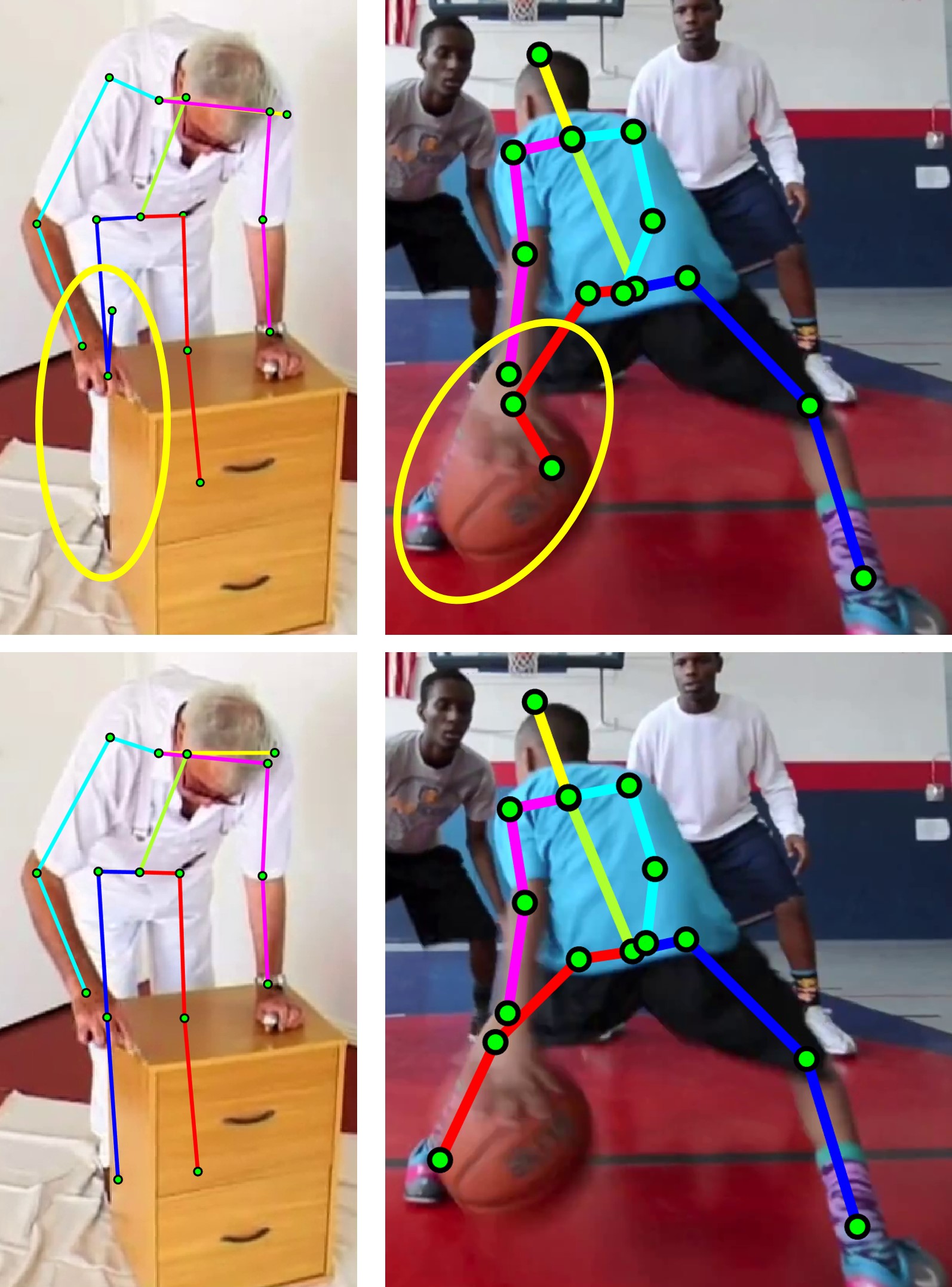}
\end{minipage}
}
\hspace{0.15cm}
\subfigure[Nearby person]{
\begin{minipage}[t]{0.15432\linewidth}
\centering
\includegraphics[width=0.925\columnwidth]{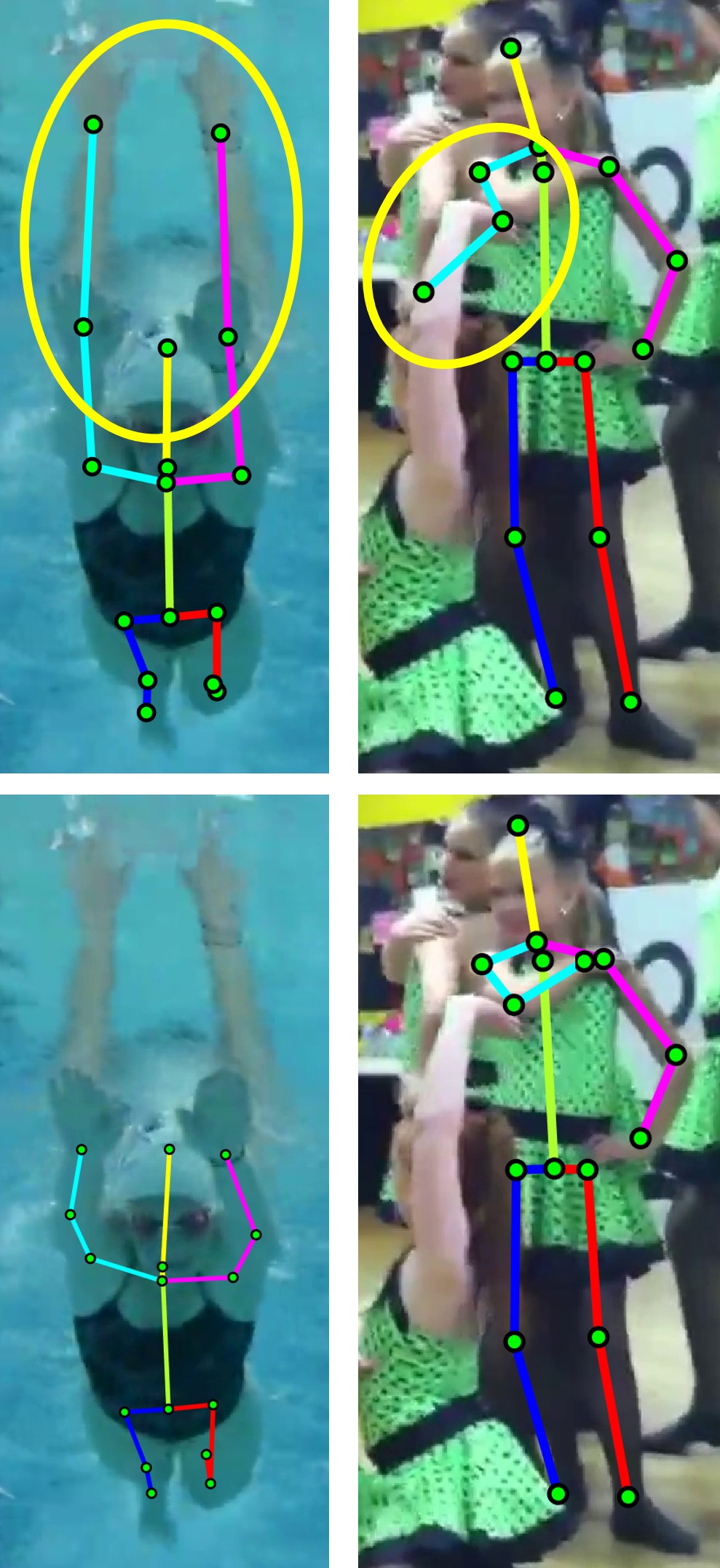}
\end{minipage}
}
\hspace{0.15cm}
\subfigure[Human symmetric similar appearance]{
\begin{minipage}[t]{0.41624\linewidth}
\centering
\includegraphics[width=1.0\columnwidth]{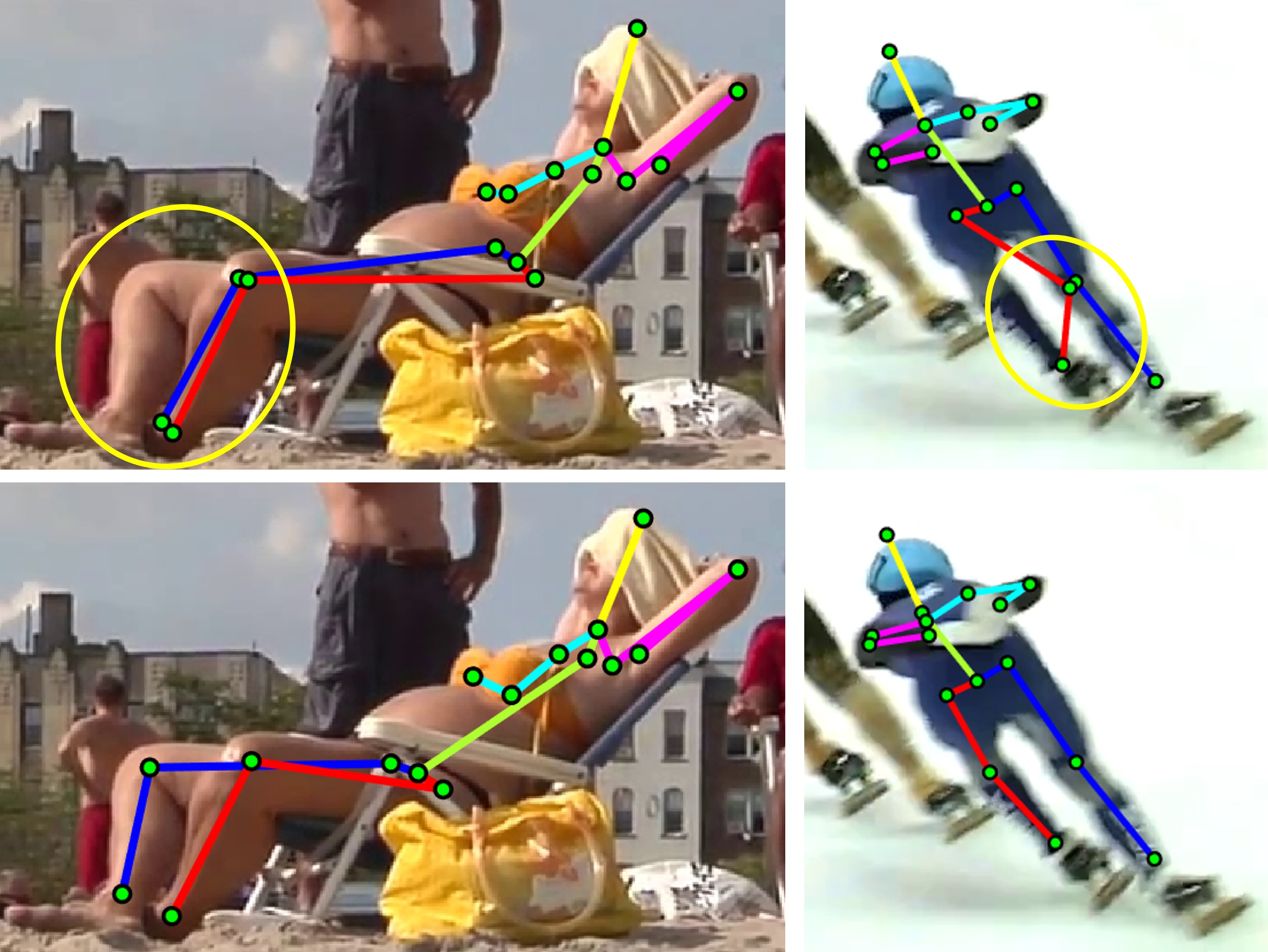}
\end{minipage}
}
\centering
\caption{\textbf{Illustration of confusing textures in human pose estimation.} 
\textbf{Top}: Pose predictions from DLCM~\cite{tang2018deeply}. 
\textbf{Bottom}: Ours. 
In case of heavy occlusion, DLCM gets error prediction on ankles, while ours successfully captures the details of white trouser and basketball shoe and give correct prediction. 
For nearby person, DLCM is confused by the surrounding arms, while ours learns to resist the interference.
Moreover, due to human's symmetric similar appearance, DLCM becomes confused between the autologous knees, while ours correctly distinguishes the symmetric joints.}
\end{figure}

\vspace{-0.25cm} 
Human Pose Estimation (HPE) localizes human anatomical keypoints (joints), which plays an important role in a variety of high-level vision tasks, such as action recognition~\cite{wang2013approach}, human tracking~\cite{xiao2018simple}, human image synthesis~\cite{ma2017pose},~\textit{etc}. 
The recent advances show that Deep Convolutional Neural Networks (DCNN) have achieved state-of-the-art performance~\cite{zhang2019human,sun2019deep,tang2018deeply,ke2018multi}. 
However, these networks can still be easily confused by three kinds of challenging cases: heavy occlusion, nearby person and symmetric appearance, which we term as \textit{confusing textures}, as shown in Fig.~\ref{fig_confusing}.

Confusing textures in heavy occlusion and nearby person are self-explanatory, which have been widely studied and addressed in~\cite{zhang2019human,ke2018multi,fieraru2018learning,tang2018deeply}.
However, symmetric appearance is resulted from the highly symmetric similarity, such as the shoes, sleeves, trousers and so on, which was rarely explored.
We experimentally observe that the symmetric human appearance can easily confuse the network, especially in case of crossed arms and legs.
To our best knowledge, it is the first work to identify the concept of symmetric appearance and discuss its influence on human pose estimation.

How to resolve the confusing textures is a core problem in human pose estimation. 
Typically, the previous works focus on three aspects to achieve anti-confusing:
First, various kinds of data augmentation schemes are adopted, including scaling, rotating, flipping, image-synthesis with Motion Capture \cite{rogez2016mocap}, joint-switch~\cite{fieraru2018learning}, keypoint-masking~\cite{ke2018multi},~\textit{etc}.
Second, effective feature learning is pivotal.
For example, the popular hourglass model~\cite{newell2016stacked} and its variants~\cite{chen2017adversarial,xiao2018simple,zhang2019human,yang2017learning,tang2018quantized,sun2019deep} stack different kinds of high-low-high sub-networks that effectively learn the high-level features for heatmap prediction.
Third, adaptive prediction fusion on multiple candidate heatmaps or coordinates~\cite{yang2017learning,zhang2019human} is a key step to improve accuracy of final prediction.
Despite the great success achieved by the above methods, none of these methods address the three aspects in a single framework simultaneously.

\begin{figure}[t!]
\setlength{\abovecaptionskip}{-0.0cm} 
\setlength{\belowcaptionskip}{-0.35cm}
\centering
\includegraphics[width=0.95\textwidth]{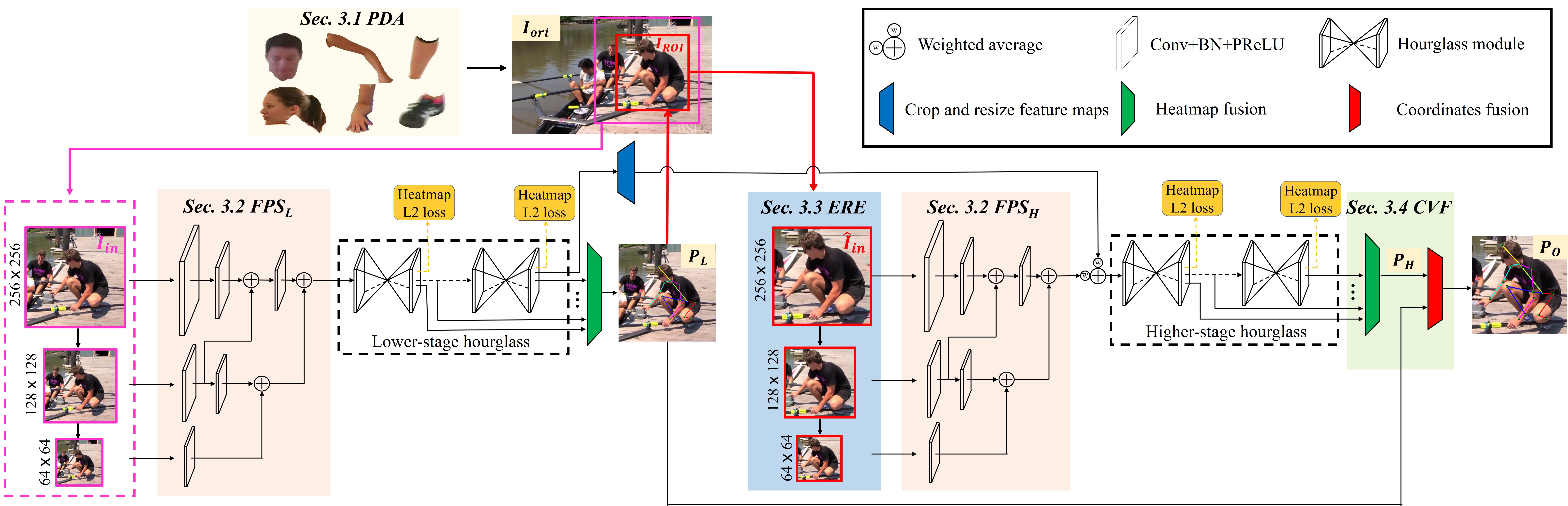}
\caption{\textbf{Framework of our Region-Aware Network.} 
The architecture is based on stacked hourglass network~\cite{newell2016stacked} with four novel modules, including Parsing-based Data Augmentation (Sec.~\ref{sec:PDA}) that exploits human semantic information; Feature Pyramid Stem (Sec.~\ref{sec:FPS}) for better low-level features and Effective Region Extraction (Sec.~\ref{sec:ERE}) for better target-specific features; and finally Cascaded Voting Fusion (Sec.~\ref{sec:CVF}) for more accurate joint prediction.
}
\label{fig_pipeline}
\end{figure}

To address the above issues, we propose a novel framework, namely Region-Aware Network (RANet), as shown in Fig.~\ref{fig_pipeline}, which comprehensively addresses the three aspects for human pose estimation.
First, to synthesize various kinds of confusing textures, we propose a \textit{Parsing-based Data Augmentation} (PDA), which segments the body parts and mounts them on or around the current human's joints.
Second, different from the previous methods that focus on learning effective high-level features, we on one hand emphasize the important role of the low-level features, and thus propose a \textit{Feature Pyramid Stem} (FPS) to excavate the input image under different resolutions; on the other hand emphasize the role of pixel region of human body, and hence propose an \textit{Effective Region Extraction} (ERE) module to crop the effective region according to the bounding box estimated from the intermediate prediction, which greatly reduces the effect of background and extract more target-specific features.
Finally, after getting multiple candidate predictions, we propose \textit{Cascade Voting Fusion} (CVF) module to explicitly exclude the inferior predictions and merge the rest ones in a weighted manner as the final output.
We evaluate our method on two representative benchmark datasets, \textit{i.e.} MPII human pose dataset (MPII)~\cite{andriluka20142d} and extended Leeds Sports Poses (LSP), and achieve state-of-the-art performance. 
Especially on easily-confusable joints, like elbow, wrist, knee and ankle, our method achieves significant improvement.
In summary, the main contributions of this work are three-folds:

$\bullet$ We identify different kinds of confusing textures, and propose a novel parsing-based augmentation scheme, which forces the network to learn the ability of anti-confusing.

$\bullet$ We propose a feature pyramid stem module to learn stronger low-level features and an effective region extraction module to extract better target-specific features.

$\bullet$ We design a cascade voting fusion module that explicitly excludes the inferior prediction and merge the rest superior predictions as the final output, which is effective and can be easily applied on other multi-prediction frameworks.

\vspace{-0.1cm}
\section{Related Work}
\vspace{-0.3cm}
Previous DCNN based human pose estimation works can be roughly divided into two categories.
The first group directly regresses the location coordinates of joints~\cite{toshev2014deeppose,yu2016deep}, called regression-based methods.
The second group predicts heatmaps followed by estimating joint locations according to the peak or integration response of heatmap~\cite{wei2016convolutional,newell2016stacked}, termed heatmap-based methods. Our work is closely related to the second group while differing from three perspectives.

\textbf{Data Augmentation \ } 
Conventional data augmentation methods on human pose estimation task~\cite{newell2016stacked,chen2017adversarial,zhang2019human} mainly performs scaling, rotating and flipping, \textit{etc} on the training images. 
Recently, PoseRefiner~\cite{fieraru2018learning} mimics incorrect pose joints and refines them by cascaded network.
MSR-net~\cite{ke2018multi} introduces keypoint-masking to simulate the hard training samples.
Different from the previous data augmentation strategies, we propose a novel parsing-based data augmentation scheme that taking advantage of the semantic segmentation for synthesizing various confusing situations.

\textbf{Feature Learning \ } 
Most previous networks~\cite{newell2016stacked,xiao2018simple,sun2019deep,zhang2019human,chen2017adversarial,chu2016crf} focus on learning effective high-level features for heatmap prediction, which incorporates the hourglass-like structure that includes down-sampling for feature encoding and up-sampling for heatmap decoding.
Typically, before entering the heatmap prediction sub-network, a rough stem module that converts the input image to smaller feature maps is adopted to reduce the complexity.
For instance, stacked hourglass~\cite{newell2016stacked} accepts input in resolution of $256\times256$ while generates feature map of $64\times64$ for heatmap prediction. 
Simple-baseline~\cite{xiao2018simple} and HRNet~\cite{sun2019deep} extract $64\times48$ low-resolution feature map for heatmap prediction from $256\times192$ high-resolution input image.
However, the rough stem module may not make full use of the effective pixel-level information from the raw input images. 
In contrast, we on one hand propose a Feature Pyramid Stem module for learning stronger low-level feature; on the other hand propose an Effective Region Extraction module for better target-specific features.

\textbf{Prediction Fusion \ } 
Prediction fusion strategy is a common solution to improve the hard joints in challenging cases.
Zhang \textit{et. al}~\cite{zhang2019human} designs a Cascade Prediction Fusion (CPF) network that takes all heatmaps in different stages into considerations for final prediction.
Yang \textit{et. al}~\cite{yang2017learning} concatenate coarse output heatmaps with raw input for further keypoints refinement.
Compared with these methods, our method explicitly excludes the inferior candidate predictions by voting and gets more accurate results by merging the rest superior predictions.

\vspace{-0.1cm}
\section{Method}
\vspace{-0.3cm}
\subsection{Parsing-based Data Augmentation} \label{sec:PDA}
\begin{figure}[t!]
\setlength{\abovecaptionskip}{-0.2cm} 
\setlength{\belowcaptionskip}{-0.3cm}
\centering
\begin{minipage}[b]{\textwidth}
\centering
\includegraphics[width=0.96\textwidth]{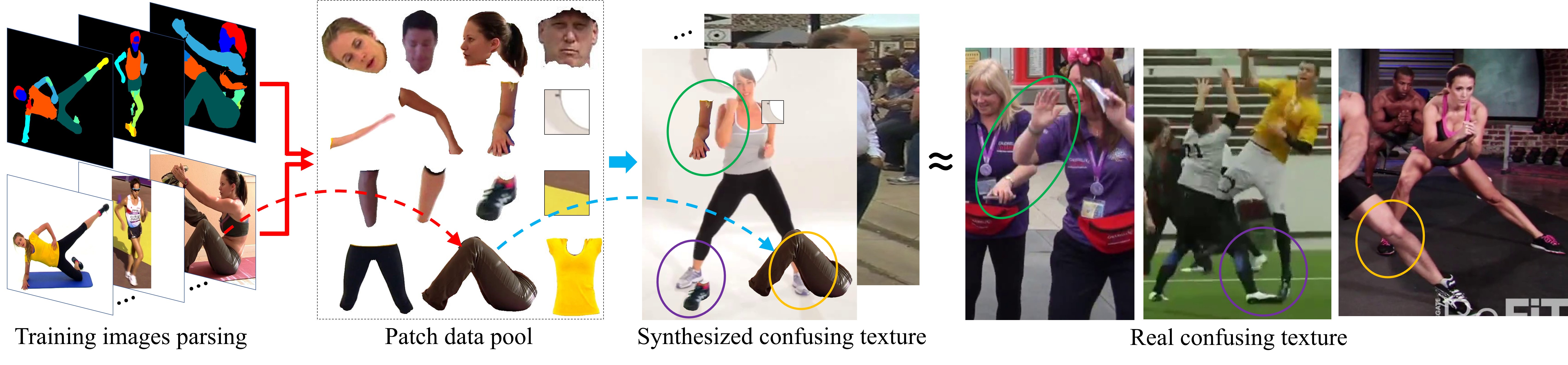} \\
\vspace{-0.2cm}
\hdashrule[0.5ex][c]{\linewidth}{0.5pt}{1.5mm} \\
\vspace{-0.15cm}
\subfigure[Joint-Switch \cite{yang2017learning}]{
\begin{minipage}[t]{0.265\linewidth}
\centering
\includegraphics[width=1.0\columnwidth]{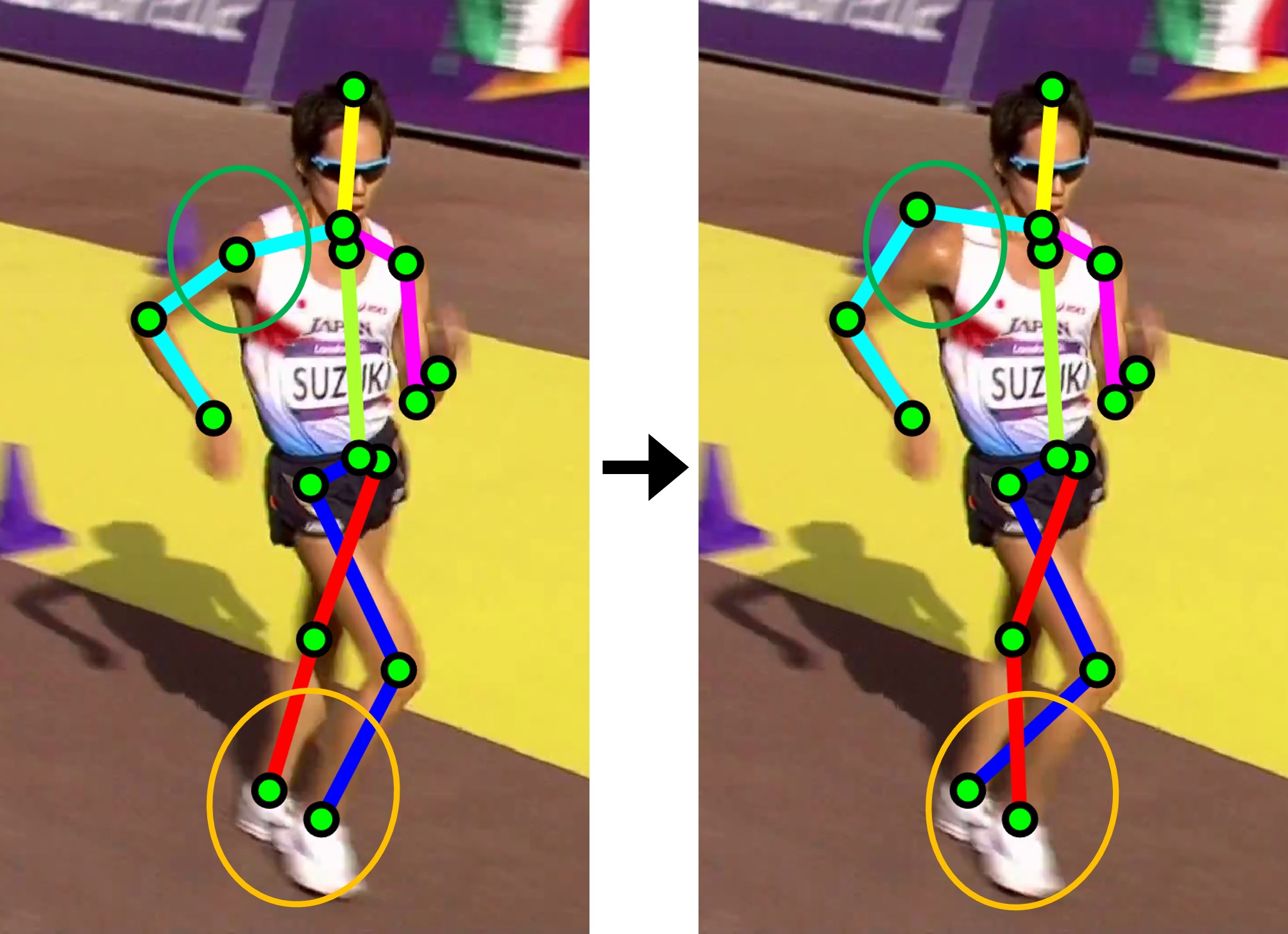}
\end{minipage}
}
\hspace{0.2cm}
\subfigure[Keypoint-Masking \cite{ke2018multi}]{
\begin{minipage}[t]{0.265\linewidth}
\centering
\includegraphics[width=1.0\columnwidth]{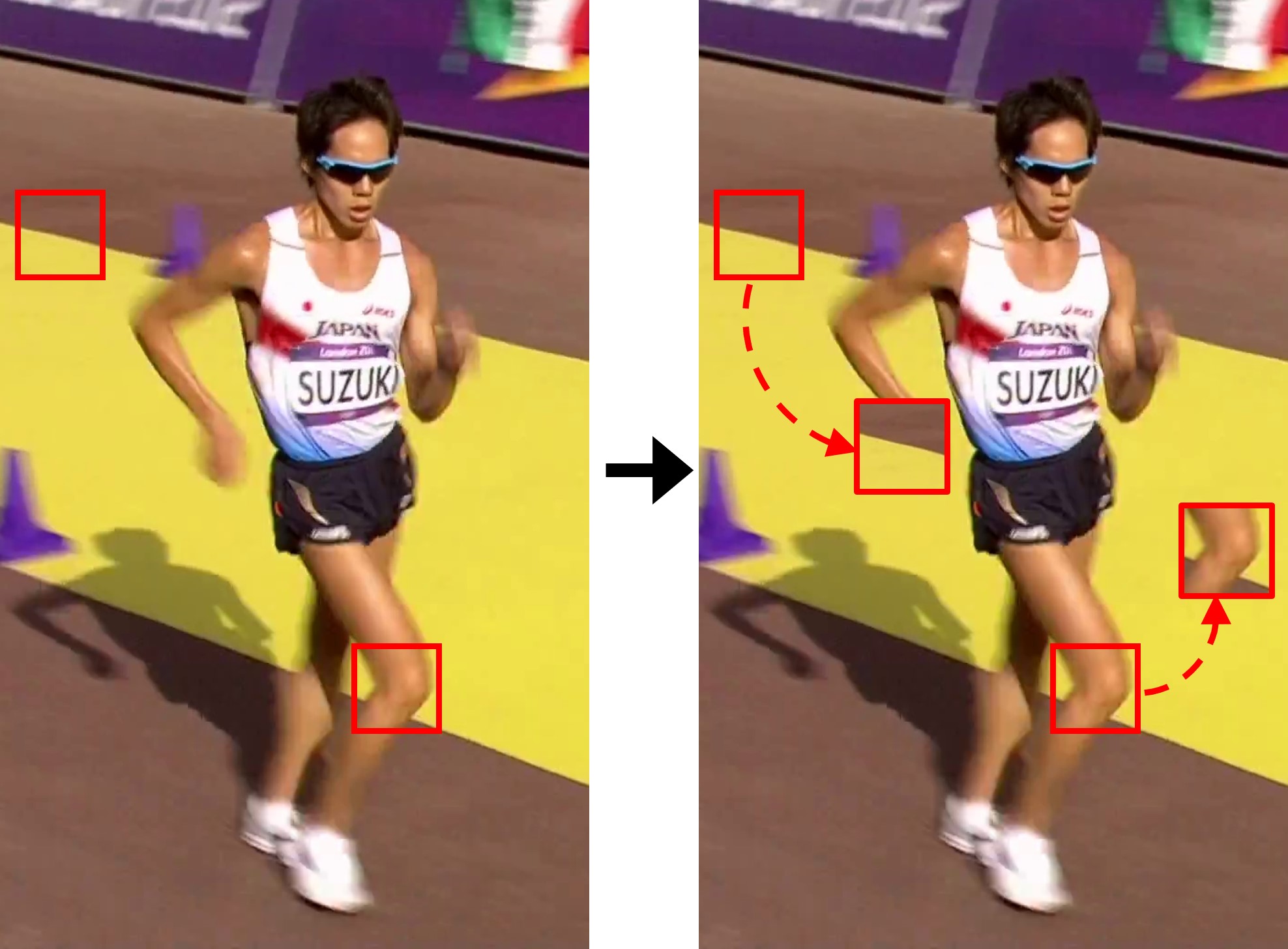}
\end{minipage}
}
\hspace{0.2cm}
\subfigure[Our Parsing-based Data Aug.]{
\begin{minipage}[t]{0.3376\linewidth}
\centering
\includegraphics[width=1.0\columnwidth]{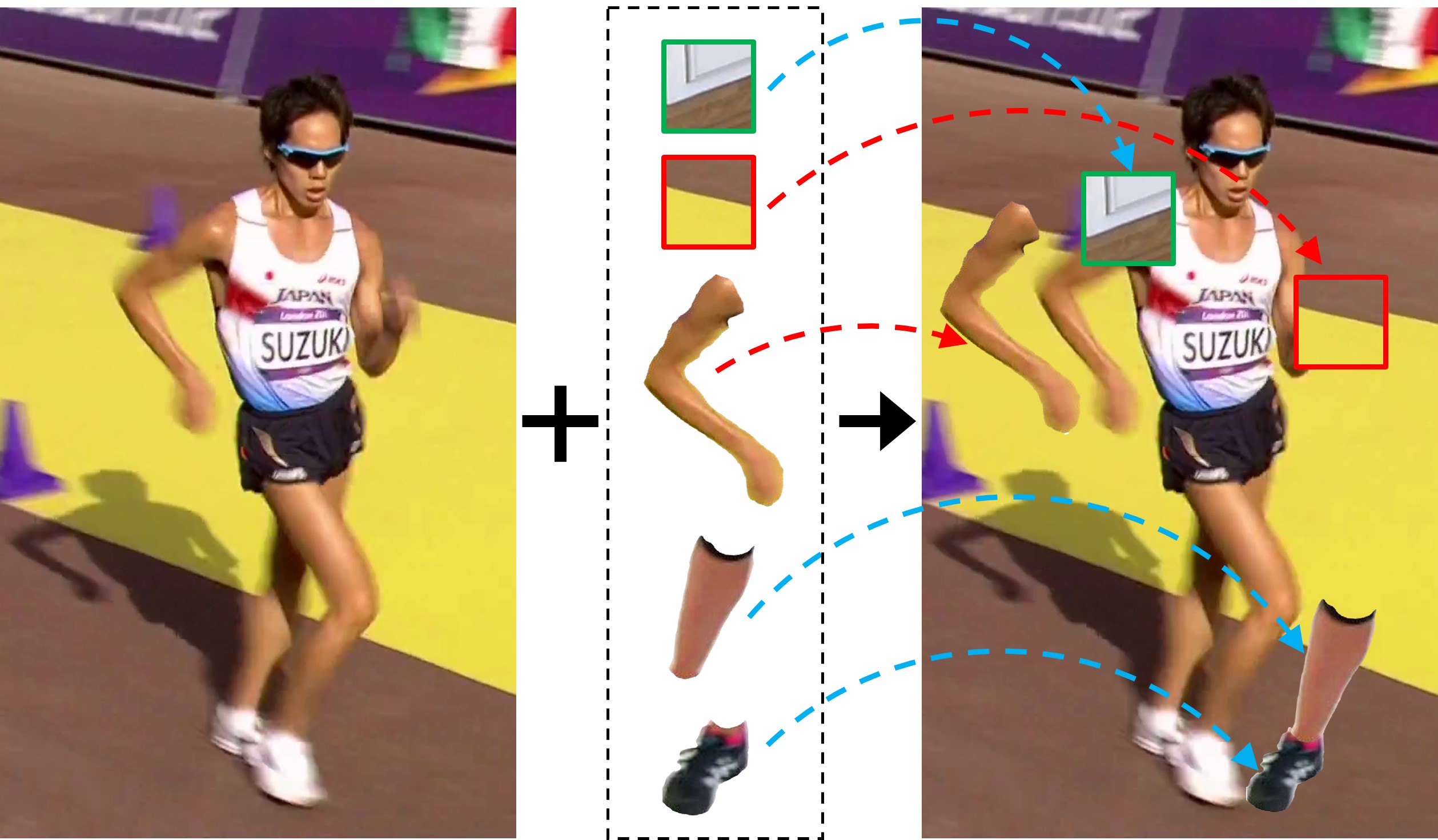}
\end{minipage}
}
\end{minipage}
\caption{\textbf{Top: Illustration of parsing-based data augmentation.}
We first apply human parsing on the training images to build a data pool filled with segmented body parts and cropped background patches, shown as the red arrow. 
Then the patches are \textit{semantically} mounted on the training data, as the blue arrow illustrated, which may occlude the joints or appear around the joints. 
As the result, our PDA synthesizes confusing textures that are similar to the real cases, such as the nearby person parts in green circle and the occlusion in yellow circle.
\textbf{Bottom: Comparisons with related works.} (a) Joint-switch from PoseRefiner~\cite{fieraru2018learning}. (b) Keypoint-masking strategy in MSA net~\cite{ke2018multi}. (c) Ours.}
\label{fig_parsing_aug}
\end{figure}

\vspace{-0.3cm}
For common human pose estimation methods~\cite{newell2016stacked, xiao2018simple, tang2018deeply, sun2019deep, zhang2019human}, data augmentations on scale, rotation, flipping are usually applied. 
However, these augmentation methods are not strong enough to provide robustness against confusing textures. 
Here, we propose a novel augmentation scheme, as shown in Fig.~\ref{fig_parsing_aug} (Top). By firstly segments all of the training images $I_{train}$ through the state-of-the-art human parsing method $\boldsymbol{\mathcal{F}_{parsing}}$~\cite{liu2018devil,gong2017look}, then we can build a data pool $\mathbb{D}_{parsing}$ filled with various semantic body parts and background patches~\cite{ke2018multi}.
Finally, the body parts or patches from the data pool are properly mounted on the current person's body to synthesize confusing textures.

As shown in Fig.~\ref{fig_parsing_aug}~(a), Pose-Refiner~\cite{yang2017learning} mimics hard joints by only manipulating keypoints' location, like shifting or switching, which ignores the importance of augmentation on region pixel level. 
Instead, our method simulates the challenging cases by synthesizing the texture in pixel level according to the semantic body parts from a large collected data pool.
MSA net~\cite{ke2018multi} proposes a keypoint-masking strategy to move the cropped image patches, as shown in Fig.~\ref{fig_parsing_aug}~(b). Compared to the keypoint-masking, except for the cropped background patch, our PDA mounts the segmented body parts with semantics.
For example, the segmented shank patch will be mounted on or around keens/ankles in much higher probability.
As a result, our training data consists of much stronger confusing texture which forces our network to learn better ability of anti-confusing.

\vspace{-0.2cm}
\subsection{Feature Pyramid Stem} \label{sec:FPS}
\vspace{-0.2cm}
Typically, feature maps in much lower resolution are passed through the heatmap prediction sub-network, such as $64\times64$ in stacked hourglass~\cite{newell2016stacked} and $64\times48$ in simple baseline~\cite{xiao2018simple}. 
In order to alleviate the information squeezing during converting the input image to the lower resolution feature map, we introduce a Feature Pyramid Stem (FPS) module to learn stronger low-level features, which includes multiple branches to extract and merge the features in different resolutions.
Specifically, our FPS repeatedly performs the downsample process on different input resolutions until reaching the target feature resolution (\textit{e.g.}, $64$). 
Partially inspired by HRNet \cite{sun2019deep}, we further add malposed paths over different branches to merge the same-resolution features.

Given input image $I_{in}$ of size $W \times H \times 3$ that is cropped from $I_{ori}$ via the ground truth center and scale parameters for a specific human body, the proposed FPS consistses of $K$ branches to extract features. 
The input resolution for the $i^{th}$ branch is $1/(2^{i-1}) \times [W, H]$, $i\in[1,K]$.
For example, the original downsample network in stacked-hourglass~\cite{newell2016stacked} uses $256\times256$ as the input resolution and outputs heatmap in resolution of $64\times64$, in which the stride $\times2$ convolutional and max pooling layers are adopted. 
To match the lowest feature map resolution $64\times64$, we set $K=3$. 
The $1st$ branch keeps the same with the original downsample process adopted in~\cite{newell2016stacked}.
The $2nd$ branch accepts input resolution of $128\times128$ followed by the similar downsample process excluding the max pooling. 
The input resolution for the $3rd$ branch is $64\times64$ and stride $\times1$ convolutional layer is applied.
Fig.~\ref{fig_pipeline} illustrates our FPS, which could also be widely applied on other feature extraction networks.

\vspace{-0.2cm}
\subsection{Effective Region Extraction} \label{sec:ERE}
\vspace{-0.2cm}
Due to the high degree of freedom in human pose configures, like squatting, up-right and lie-low, the limbs could reach to different scopes. 
Supposing $I_{in}$ to be with resolution of $256\times256$, the full body may occupy different region sizes. 
In some cases, the pixel region of body is very small compared to the input resolution, which makes the network to pay lots of attention on the useless background regions.
For top-down human pose estimation~\cite{papandreou2017towards,jin2017towards}, an extra detector is applied at first, and the previous work~\cite{fang2017rmpe} has shown that the subtle deviation of bounding box could result in serious errors in pose estimation. 
Motivated by~\cite{fang2017rmpe}, we adopt the intermediate prediction $P_L$, from the lower stage hourglass module $\mathbf{HG_{L}}$, as a free detector to provide a more accurate and compact bounding box for the human body, which further helps the network to focus on the more effective human body region $I_{ROI}$ for better target-specific features.
To keep the aspect ratio of cropped image, we resize the $I_{ROI}$ until the long side matches to $256$ and then set the resized $I_{ROI}$ as $\hat{I}_{in}$.
As the blue-masked module shown in Fig.~\ref{fig_pipeline}, $\hat{I}_{in}$ passes through $\mathbf{FPS_{H}}$ module. 
The feature maps $F_{L}^{HG}$ and $F_{H}^{64}$ from $\mathbf{HG_{L}}$ and $\mathbf{FPS_{H}}$ are then merged as the input for the following higher-stage hourglass $\mathbf{HG_{H}}$, where $64$ indicates the resolution of feature map. It should be noted that the two FPS modules have exactly the same network structure but different parameters. Please refer to the Tab. \ref{tab_algorithm} for more details.

\begin{table}[t!]
  \setlength{\belowcaptionskip}{0.05cm}
  \small
  \caption{The data flow of algorithm}
  \label{tab_algorithm}
  \centering
  \begin{tabular}{r|ll}
    \toprule[1.2pt] 
    \multicolumn{3}{l}{\  \textbf{\# - - - Apply Parsing-based Data Augmentation}} \\
    1:& \textbf{if} training == True \\
    2:& \qquad $\mathbb{D}_{parsing} = \boldsymbol{\mathcal{F}_{parsing}}(I_{train})$ & // Build patch data pool \\
    3:& \qquad $I_{ori}=\boldsymbol{\mathcal{F}_{aug}}(I_{train}, \mathbb{D}_{parsing})$ & // Augment training data \\
    4:& \textbf{if} testing == True \\
    5:& \qquad $I_{ori}=I_{test}$ \\
    \multicolumn{3}{l}{\  \textbf{\# - - - Process flow in lower-stage hourglass sub-network}} \\
    6:& $I_{in}^{256}=\boldsymbol{\mathcal{F}_{crop}}(I_{ori}, Center, Scale)$ & // Crop image by center and scale \\
    7:& $F_{L}^{64}=\mathbf{FPS_{L}}(I_{in}^{256}, I_{in}^{128}, I_{in}^{64})$ & // Get low-level feature \\
    8:& $H_{L}, F_{L}^{HG}=\mathbf{HG_{L}}(F_{L}^{64})$ & // Get heatmap and feature \\
    9:& $P_{L},V_{L}=\boldsymbol{\mathcal{F}_{Peak}}(\mathbf{CVF_{Heatmap}}(H_{L}))$ & // Get prediction and visibility \\
    10:& $Bbox=\boldsymbol{\mathcal{F}_{bounding}}(P_{L})$ & // Get bounding box \\
    11:& $\hat{I}_{in}^{256}=\boldsymbol{\mathcal{F}_{crop}}(I_{ori}, Bbox)$ & // Crop image by bounding box \\
    \multicolumn{3}{l}{\  \textbf{\# - - - Process flow in higher-stage hourglass sub-network}} \\
    12:& $F_{H}^{64}=\mathbf{FPS_{H}}(\hat{I}_{in}^{256}, \hat{I}_{in}^{128}, \hat{I}_{in}^{64})$ & // Get low-level feature \\
    13:& $H_{H}=\mathbf{HG_{H}}(F_{H}^{64},F_{L}^{HG})$ & // Get heatmaps of higher-stage HG \\
    14:& $P_{H},V_{H}=\boldsymbol{\mathcal{F}_{Peak}}(\mathbf{CVF_{Heatmap}}(H_{H}))$ & // Get prediction and visibility\\
    \multicolumn{3}{l}{\  \textbf{\# - - - Get the final output prediction by fusing all candidates}} \\
    15:& $\{P_{L}^{i},V_{L}^{i},P_{H}^{i},V_{H}^{i}\}=\boldsymbol{\mathcal{F}_{Infer}}(I_{ori}, Scale^{i}), i\in[1,6]$ &  // Multiple inference in 6 scales\\
    16:& $P_{O}=\mathbf{CVF_{Coordinate}}(\{P_{L}^{i},V_{L}^{i},P_{H}^{i},V_{H}^{i}\})$ & // Coordinate fusion\\
    \bottomrule[1.2pt]
  \end{tabular}
\vspace{-0.4cm}
\end{table}

\vspace{-0.1cm}
\subsection{Cascade Voting Fusion} \label{sec:CVF}
\begin{figure}[H]
\setlength{\abovecaptionskip}{-0.05cm}
\centering
\includegraphics[width=0.45\textwidth]{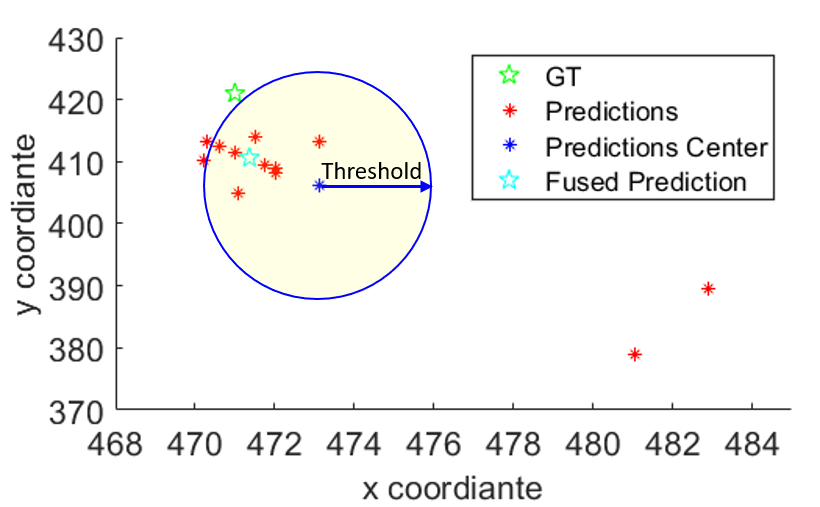}
\caption{Illustration of voting fusion}
\label{fig_fusion}
\end{figure}

\vspace{-0.2cm}
After the feature learning module, multiple predictions are estimated, from either multiple inferences in different scales~\cite{chu2017multi,yang2017learning,tang2018deeply,sun2019deep,zhang2019human} or different stages~\cite{newell2016stacked}.
How to adaptively fuse these predictions is a key step for improving the accuracy of joints.
Compared to the fusion method in~\cite{zhang2019human}, our proposed Cascade Voting Fusion explicitly excludes the inferior predictions and then fuses the rest superior predictions.
Specifically, our CVF firstly performs fusion on heatmaps and then coordinates. 
As shown in Fig.~\ref{fig_pipeline}, both lower- and higher-stage hourglasses output multiple heatmap predictions, denoted as $H^{L}$ and $H^{H}$. 
During heatmap fusion, we fuse all heatmaps from each hourglass stage according to their weighted summation.
Then we calculate the corresponding coordinate from the fused heatmap as $(x,y)_{joint} = \frac{3}{4}\times(x,y)_{peak} + \frac{1}{4}\times(x,y)_{secondary}$, and denote the peak value of heatmap as $v$, where $(x,y)_{peak}$ and $(x,y)_{secondary}$ are the coordinates of the maximal and secondary value in heatmap, respectively.
During coordinate fusion, regarding to single body joint, \textit{e.g.} left wrist, supposing there are totally $N$ candidate joint predictions, as shown in Fig.~\ref{fig_fusion}. 
Then, we get the center of all candidate joints, and utilize the average distance from candidate joints to the center as $threshold = \frac{1}{N}\sum_{i=1}^{N}(\left \| (x,y)_{pre} - (x,y)_{center} \right \|_{L_2})$, which is further used for filtering the inferior joints, \textit{i.e.}, we only keep $M$ joints whose distances to the center are smaller than the threshold:
\vspace{-0.5mm}
\begin{equation}
\label{eq_fusion}
(x,y)_{output} = \frac{1}{M}\sum_{i=1}^{M}(w_i\times(x_i,y_i)_{joint}), \quad s.t.\  distance_i < threshold \\
\end{equation}
\vspace{-0.5mm}
where $(x_i,y_i)_{joint}$ is the coordinate of $i$-th candidate, $(x,y)_{output}$ represents the final joint prediction, and $w_i = {v_i}/{\sum_{i=1}^{M}(v_i)}$. 
To get the complete joint predictions, the CVF is independently performed on each body joint in parallel.
It should also be noted that our fusion strategy can be easily built on most multi-stage pose estimation frameworks.

\vspace{-0.1cm}
\subsection{Training Loss}
\vspace{-0.2cm}
The training loss is formulated as: $\left \| H_{pre} - H_{gt} \right \|_{L_2}$. 
The ground truth heatmap $H_{gt}$ is generated according to the Gaussian distribution as
$H_{(x,y)}=e^{-{(x-\hat{x})^2 \times (y-\hat{y})^2}/{2\sigma^2}}$,
where $(\hat{x},\hat{y})$ is the coordinate of the ground truth joint, and $\sigma$ represents the standard deviation that controls the energy distribution. 
Generally principle energy (about $99.73$\%) is distributed in [-$3\sigma$, +$3\sigma$]. 
Obviously, larger $\sigma$ results in the larger radius of principle energy, which may let the network more easily learn the heatmap~\cite{pfister2015flowing}. 
We add supervision with visibility by classifying joints into three categories: visible, occluded and outer. 
For visible joints, we set the heatmap with $\sigma=1$ as supervision. 
Occluded joints are more difficult to learn. 
In order to easily learn such hard joints, we set twice value $\sigma=2$ for the occluded joints. 
Finally, for outer joints, heatmap of all zeros is provided as supervision which instructs network to pay no attention on outer joints. 
By giving different kinds of heatmap supervision, the network pays different degrees of attention on learning the above three kinds of joints.

\vspace{-0.1cm}
\section{Experiments}
\vspace{-0.3cm}
\textbf{Datasets and Data Augmentation \ }
We evaluate our method on two representative benchmark datasets including MPII human pose dataset (MPII)~\cite{andriluka20142d} and extended Leeds Sports Poses (LSP)~\cite{johnson2010clustered}. 
MPII consists of $25,000$ images with over $40,000$ annotated poses. We split training and validation sets following~\cite{newell2016stacked}. 
The extended LSP dataset consists of $11,000$ training images from sport activities and $1,000$ images for testing.
We augment the training data by randomly scaling in [$0.75$, $1.25$], rotation in [$-60$, $+60$] degree, horizontal flipping and color adjustment. As mentioned in section 3.1, we further apply Parsing-based Data Augmentation on the training image. For the wrong parsing results, we simply remove the misshapen parts in manual. 

\textbf{Implementation Details \ } The network is implemented on PyTorch with optimizer RMSProp. We train the network in $250$ epochs and batch size is $32$. 
The learning rate starts at $0.0005$ and decreases by $2$ times at $20^{th}$, $50^{th}$, $100^{th}$, $150^{th}$, $200^{th}$ epoch, respectively. 
We follow the common evaluation criteria, \textit{i.e.}, Percentage Correct Keypoints (PCK) is utilized to evaluate results on LSP~\cite{zhang2019human,yang2017learning}, and PCKh~\cite{andriluka20142d} that normalizes the distance errors with respect to the size of head is leveraged for MPII. 
The input image $I_{in}$ is cropped according to the approximate human center and scale, and warped to size $256\times256$. 
The input image $\hat{I}_{in}$ is cropped according to the bounding box of ERE module. 
Following \cite{chu2017multi,yang2017learning,tang2018deeply,sun2019deep,zhang2019human}, a testing procedure that adopts six different scales is performed with horizontal flipping. 
The CVF module merges the all predictions in six scales as the final output.

\begin{table}[t!]
  \setlength{\belowcaptionskip}{0.05cm}
  \caption{Performance comparisons on the MPII test set (PCKh@0.5)}
  \label{tab_MPII_Test}
  \small
  \centering
  \begin{tabular}{p{45pt} p{50pt} p{35pt}|ccccccc|c}
    \hline
    \multicolumn{3}{c}{Method (\%)} & Head & Sho. & Elb. & Wri. & Hip & Knee & Ank. & Total \\
    \hline
    Wei \textit{et al.} & CPM \cite{wei2016convolutional} & CVPR'$16$ & 97.8  & 95.0  & 88.7  & 84.0  & 88.4  & 82.8 & 79.4 & 88.5 \\
    Bulat \textit{et al.} & PHR \cite{bulat2016human} & ECCV'$16$ & 97.9  & 95.1  & 89.9  & 85.3  & 89.4  & 85.7 & 81.7 & 89.7 \\
    Newell \textit{et al.} & Hourglass~\cite{newell2016stacked} & ECCV'$16$ & 98.2  & 96.3  & 91.2  & 87.1  & 90.1  & 87.4 & 83.6 & 90.9 \\
    Ning \textit{et al.} & KG-DFN \cite{ning2018knowledge} & TMM'$17$ & 98.1  & 96.3  & 92.2  & 87.8  & 90.6  & 87.6 & 82.7 & 91.2 \\
    Chu \textit{et al.} & HRUs \cite{chu2017multi} & CVPR'$17$ & 98.5  & 96.3  & 91.9  & 88.1  & 90.6  & 88.0 & 85.0 & 91.5 \\
    Chen \textit{et al.} & Adver-PN \cite{chen2017adversarial} & ICCV'$17$ & 98.1  & 96.5  & 92.5  & 88.5  & 90.2  & 89.6 & 86.0 & 91.9 \\
    Yang \textit{et al.} & PRMs \cite{yang2017learning} & ICCV'$17$ & 98.5  & 96.7  & 92.5  & 88.7  & 91.1  & 88.6 & 86.0 & 92.0 \\
    Xiao \textit{et al.} & SimpleBase~\cite{xiao2018simple} & ECCV'$18$ & 98.5  & 96.6  & 91.9  & 87.6  & 91.1  & 88.1 & 84.1 & 91.5 \\
    Ke \textit{et al.} & MSR-net~\cite{ke2018multi} & ECCV'$18$ & 98.5  & 96.8  & 92.7  & 88.4  & 90.6  & 89.4 & 86.3 & 92.1 \\
    Nie \textit{et al.} & PIL~\cite{nie2018human} & CVPR'$18$ & 98.6  & 96.9  & 93.0  & 89.1  & 91.7  & 89.0 & 86.2 & 92.4 \\ 
    Tang \textit{et al.} & DLCM~\cite{tang2018deeply} & ECCV'$18$ & 98.4  & 96.9  & 92.6  & 88.7  & 91.8  & 89.4 & 86.2 & 92.3 \\
    Sun \textit{et al.} & HRNet~\cite{sun2019deep} & CVPR'$19$ & 98.6  & 96.9  & 92.8  & 89.0 & 91.5  & 89.0 & 85.7 & 92.3 \\
    Zhang \textit{et al.} & PGNN \cite{zhang2019human} & arXiv'$19$ & \bf{98.6} & 97.0 & 92.8 & 88.8 & 91.7 & 89.8 & 86.6 & 92.5 \\
    \hline
    Ours & RANet & & 98.5 & \bf{97.0}  & \bf{93.4}  & \bf{89.8}  & \bf{92.0}  & \bf{90.3}  & \bf{87.6} & \bf{92.9}  \\
    \hline
  \end{tabular}
\end{table}

\begin{table}[t!]
  \setlength{\belowcaptionskip}{0.05cm}
  \caption{Performance comparisons on the LSP test set (PCK@0.2)}
  \label{tab_LSP_Test}
  \small
  \centering
  \begin{tabular}{p{45pt} p{50pt} p{35pt}|ccccccc|c}
    \hline
    \multicolumn{3}{c}{Method (\%)} & Head & Sho. & Elb. & Wri. & Hip & Knee & Ank. & Total \\
    \hline
    Insafu. \textit{et al.} & Deepercut \cite{insafutdinov2016deepercut} & ECCV'$16$ & 97.4 & 92.7 & 87.5 & 84.4 & 91.5 & 89.9 & 87.2 & 90.1 \\
    Wei \textit{et al.} & CPM \cite{wei2016convolutional} & CVPR'$16$ & 97.8 & 92.5 & 87.0 & 83.9 & 91.5 & 90.8 & 89.9 & 90.5 \\
    Bulat \textit{et al.} & PHR \cite{bulat2016human} & ECCV'$16$ & 97.2 & 92.1 & 88.1 & 85.2 & 92.2 & 91.4 & 88.7 & 90.7 \\
    Chu \textit{et al.} & HRUs \cite{chu2017multi} & CVPR'$17$ & 98.1 & 93.7 & 89.3 & 86.9 & 93.4 & 94.0 & 92.5 & 92.6 \\
    Chen \textit{et al.} & Adver-PN \cite{chen2017adversarial} & ICCV'$17$ & 98.5 & 94.0 & 89.8 & 87.5 & 93.9 & 94.1 & 93.0 & 93.1 \\
    Yang \textit{et al.} & PRMs \cite{yang2017learning} & ICCV'$17$ & 98.3  & 94.5  & 92.2  & 88.9  & 94.4  & 95.0 & 93.7 & 93.9 \\
    Zhang \textit{et al.} & PGNN \cite{zhang2019human} & arXiv'$19$ & 98.4 & 94.8 & 92.0 & 89.4 & 94.4 & 94.8 & 93.8 & 94.0 \\
    \hline
    Ours & RANet & &  \bf{98.5} & \bf{95.5}  & \bf{93.8}  & \bf{90.5}  & \bf{95.1}  &  \bf{95.2}  & \bf{94.5} & \bf{94.7}  \\
    \hline
  \end{tabular}
\end{table}

\vspace{-0.1cm}
\subsection{Comparisons with State-of-the-Art Methods}
\vspace{-0.2cm}
\textbf{Accuracy \ }
We submit our results on test set to MPII website\footnote{\url{http://human-pose.mpi-inf.mpg.de/\#evaluation}} and get the official evaluation shown in Tab.~\ref{tab_MPII_Test}. 
Our method achieves $92.9$\% on PCKh@$0.5$, which is the highest performance on average. 
It is noteworthy that, for the joints on easily-confusable body parts, including elbow, wrist, knee and ankle, our method achieves significant improvement compared to the state of the arts. It proves the effectiveness on resolving confusing texture resulted from symmetric appearance. Tab.~\ref{tab_LSP_Test} shows our results on LSP test set. 
Following previous methods~\cite{chu2017multi,yang2017learning}, we add the MPII training set to the extended LSP training set. 
Our method again outperforms the state-of-the-art methods and still maintains competitive advantages on some hard joints.

\begin{figure}[t!]
\setlength{\abovecaptionskip}{-0.05cm}
\setlength{\belowcaptionskip}{-0.35cm}
\centering
\includegraphics[width=0.9\textwidth]{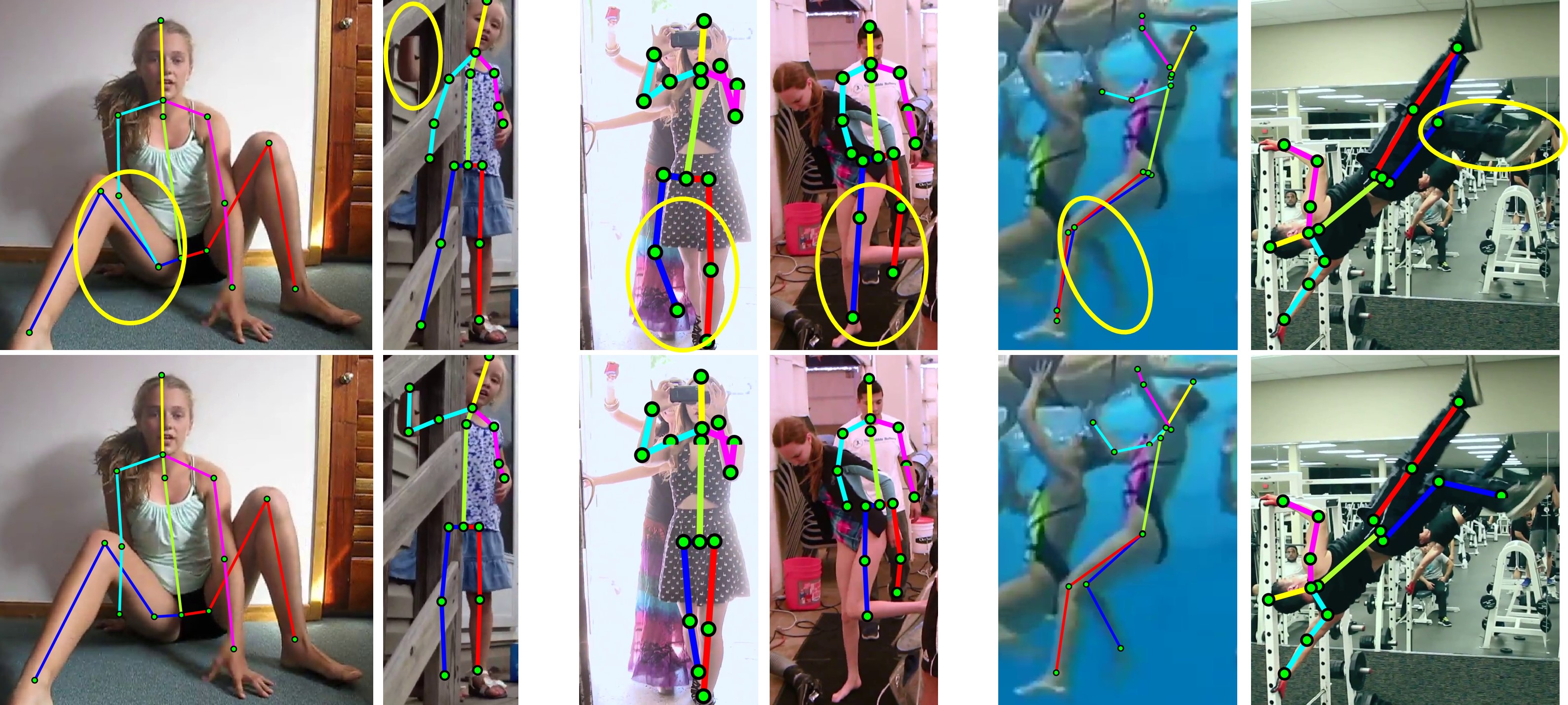}
\caption{\textbf{Qualitative results in the MPII dataset}. \textbf{Top}: DLCM~\cite{tang2018deeply}. 
\textbf{Bottom}: Ours.}
\label{fig_more_result}
\end{figure}

\textbf{Visualize \ }
In Fig.~\ref{fig_more_result}, we visualize pose estimation of DLCM~\footnote{\url{http://www.ece.northwestern.edu/~wtt450/project/ECCV18_DLCM/}}~\cite{tang2018deeply} and our model. 
The DLCM model is confused by heavy occlusion, nearby person and symmetric appearance. 
Our model successfully resolves such confusing texture in the three challenging situations. 
As shown in~\textit{cols.} $1$, the wrists are occluded and very limited hand/arm textures are exposed. 
The proposed Effective Region Extraction module helps the network to see more details, and makes it possible for the following modules to learn the looming cues. 
Interfered by the nearby person, as shown in \textit{cols.} $2$, the DLCM model predicts ankles on other person incorrectly, while our Parsing-based Data Augmentation forces the network to learn the ability of anti-confusing for nearby body parts. In addition, the proposed Cascade Voting Fusion module excludes the incorrect prediction to some extent and improves the final prediction. 
When the symmetric similarity meets image degradation or cluttered background, in \textit{cols.} $3$, it's more difficult to distinguish the symmetric joints. In such cases, our Feature Pyramid Stem module learns stronger low-level feature from the input image and provides more useful information for final heatmap prediction. 
More results are available in supplementary material. 

\vspace{-0.1cm}

\vspace{-0.1cm}
\subsection{Ablation Study}
\vspace{-0.2cm}
To investigate the effectiveness of each modules in RANet, we conduct ablative analysis on the validation set of MPII dataset. 
We adopt stacked-hourglass as backbone and achieve PCKh@$0.5$ with $88.97$\% by performing Multi-Scale and Flipping (M.S.F.) in inference as the method (a) in Tab.~\ref{tab_ablation_study}.

\textbf{PDA \ } To evaluate the contribution of Parsing-based Data Augmentation, we compare the result without and with PDA, as the methods (a) and (b). 
The PDA brings significant improvement on accuracy by 1.14\%. Although a little manual labour is necessary to remove misshapen parsing results, such one-off cost is worthy to the significant improvement. And this scheme could be widely adopted by other similar tasks.

\textbf{FPS \ } Compared to the single branch in original stacked-hourglass network~\cite{newell2016stacked}, we further add Feature Pyramid Stem module that contains extra branches to learn low-level features in different resolutions, as method (c). 
The stronger low-level feature yields $0.35$\% improvement. It's noteworthy that our FPS module add only $1.96$\% extra parameters compared with original hourglass~\cite{newell2016stacked}.

\textbf{ERE \ } We compare performance without and with Effective Region Extraction module, as methods (c) and (d). 
Although the individual ERE module brings modest improvement with $0.23$\%, it is very helpful for the following fusion and it doesn't increase network parameters.

\begin{figure}[H]
\setlength{\abovecaptionskip}{-0.05cm}
\centering
\includegraphics[width=0.45\textwidth]{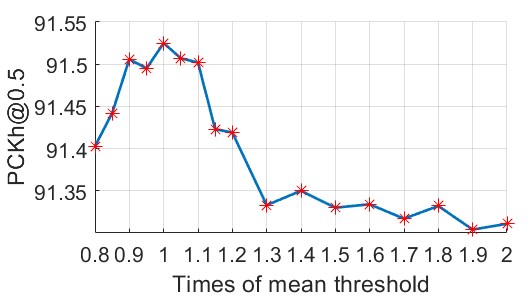}
\caption{Performance under different thresholds}
\label{fig_differentThres}
\end{figure}

\textbf{CVF \ } To study the effectiveness of Cascade Voting Fusion module, we conduct two experiments as methods (e) and (f). 
In (e), we add CVF on basis of (c) excluding ERE module, and the CVF makes $0.6$\% increment on accuracy. 
In (f), the ERE is included, so the CVF fuses more candidate predictions. 
Consequently we achieve $1.06$\% improvement by adding both ERE and CVF.
In addition, we study the influence of different $threshold$ on the pose prediction accuracy, as shown in Fig. \ref{fig_differentThres}. We denote the average distance from candidate to their center as mean threshold. Larger or smaller thresholds than the mean threshold bring worse results. For a extreme large threshold which covers all candidates, which means the inferior candidates are not be excluded, we achieve PCKh@0.5 with 91.24 which drops 0.28\% from that in mean threshold.

\vspace{-0.1cm}
\begin{multicols}{2}
\begin{table}[H]
  \setlength{\belowcaptionskip}{0.05cm}
  \caption{Ablative analysis on MPII validation set}
  \label{tab_ablation_study}
  \small
  \centering
  \begin{tabular}{c|c|c|c|c|c|c}
    \hline
      & M.S.F & PDA & FPS & ERE & CVF & PCKh \\
    \hline
    a & \checkmark &  &  &  &  & 88.97 \\
    b & \checkmark & \checkmark &   &   &   & 90.11 \\
    c & \checkmark & \checkmark & \checkmark &  &  & 90.46 \\
    d & \checkmark & \checkmark & \checkmark & \checkmark &  & 90.69 \\
    e & \checkmark & \checkmark & \checkmark &  & \checkmark & 91.06 \\
    f & \checkmark & \checkmark & \checkmark & \checkmark & \checkmark & 91.52 \\
    \hline
  \end{tabular}
\end{table}
\vspace{-0.7cm}

\vspace{-0.5cm}

\vspace{-0.4cm}

\begin{figure}[H]
\setlength{\abovecaptionskip}{0.05cm}
\centering
\includegraphics[width=0.48\textwidth]{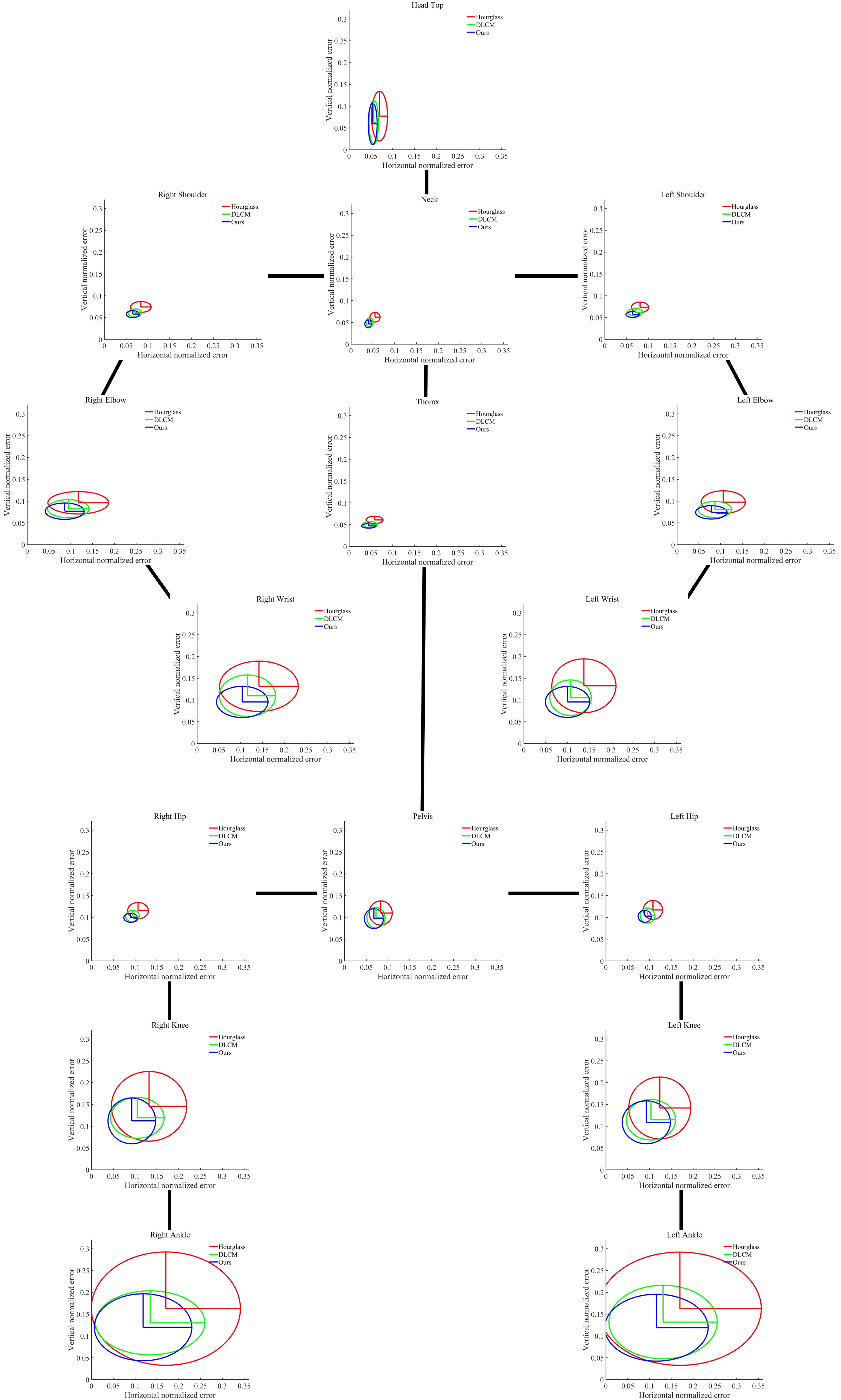}
\caption{Statistic analysis on each body joint}
\label{fig_mean_std}
\end{figure}
\end{multicols}
\vspace{-0.4cm}

\textbf{Statistic Analysis \ } We further study the statistic distribution of prediction errors among hourglass~\cite{newell2016stacked}, DLCM~\cite{tang2018deeply} and ours, as shown in Fig.~\ref{fig_mean_std}. 
We utilize the size of head as the normalized distance which is the same with that in calculation of PCKh~\cite{andriluka20142d}.
Specifically $d_{norm}=0.6\times\sqrt{W_{bboxh}^2 + H_{bboxh}^2}$, where $W_{bboxh}$ and $H_{bboxh}$ are the width and height of head's bounding box respectively. 
We calculate the horizontal and vertical errors of joints by absolute value: $x_{err} = \frac{\left \| x_{pre} - x_{gt} \right \|_{L1}}{d_{norm}}$, $y_{err} = \frac{\left \| x_{pre} - x_{gt} \right \|_{L1}}{d_{norm}}$. 
The center and radius of ellipse represent the mean and variance of prediction errors. 
Based on quantitative comparison, our method achieves smaller mean error over all joints and variance on most joints than DLCM \cite{tang2018deeply}. 
Moreover, there are some interesting conclusions: 
($1$) The joints of limbs generally have larger mean and variance of errors, especially the wrists and ankles, which proves higher difficulty degree of predicting such "hard joints". 
($2$) Most joints have larger variance in horizontal coordinate error which indicates that it's more difficult for prediction in horizontal direction. 
However, the joint of head top has much larger variance error in vertical direction. 
($3$) On every joint, the advances of methods steady move the center close to the zero, but the three methods still share very similar ellipse shape.

\vspace{-0.1cm}
\section{Conclusion}
\vspace{-0.3cm}
In this work, we propose Region-Aware Network (RANet) to effectively resolve the confusing texture in single person pose estimation. Experimental results have demonstrated the effectiveness of our approach. The success stems from Parsing-based Data Augmentation scheme and three novel modules, \textit{i.e.}, Feature Pyramid Stem (FPS), Effective Region Extraction (ERE) and Cascade Voting Fusion (CVF). FPS reinforce feature learning in lower stage which provides more useful information for the following heatmap prediction. ERE detect human body for free and extract the uttermost pixel region which help the network to see more details. CVF exclude the inferior predictions and adaptively fuse the rest superior prediction for more accurate pose estimation. In the future, we plan to extend our works on multiple person pose estimation and pose tracking in frame sequence.
\medskip

{\small
\bibliographystyle{plain}
\bibliography{ref}
}

\end{document}